\documentclass[11pt]{article}
\usepackage{fullpage}
\date{}
\pdfoutput=1

\usepackage{url}  
\usepackage{graphicx}  
\usepackage{graphicx, amsmath, amssymb}
\usepackage{subfig}
\usepackage{color}
\usepackage{array}

\usepackage{enumerate}

\DeclareMathOperator*{\argmin}{argmin}
\DeclareMathOperator*{\argmax}{argmax}

\usepackage[ruled,vlined]{algorithm2e}
\ifx \DontPrintSemicolon \undefined
 \def \DontPrintSemicolon {\dontprintsemicolon}
\fi

\graphicspath{{./images/}}

\usepackage[normalem]{ulem}
\usepackage[utf8]{inputenc}
\usepackage[dvipsnames]{xcolor}

\begin{document}

\title{\LARGE \bf Multi-UAV Visual Coverage of Partially Known 3D Surfaces: Voronoi-based Initialization to Improve Local Optimizers}
\author{Alessandro Renzaglia, Jilles Dibangoye, Vincent Le Doze and Olivier Simonin
\thanks{Authors are with INRIA Grenoble and INSA Lyon CITI Lab, Chroma team, France;
e-mail: {\tt\small firstname.lastname@inria.fr}}}

\maketitle
\thispagestyle{empty}
\pagestyle{empty}

\begin{abstract}
In this paper we study the problem of steering a team of Unmanned Aerial Vehicles (UAVs) toward a static configuration which maximizes the visibility of a 3D environment. The UAVs are assumed to be equipped with visual sensors constrained by a maximum sensing range and the prior knowledge on the environment is considered to be very sparse. To solve this problem on-line, derivative-free measurement-based optimization algorithms can be adopted, even though they are strongly limited by local optimality. To overcome this limitation, we propose to exploit the partial initial knowledge on the environment to find suitable initial configurations from which the agents start the local optimization. In particular, a constrained centroidal Voronoi tessellation on a coarse approximation of the surface to cover is proposed. The behavior of the agent is so based on a two-step optimization approach, where a stochastic optimization algorithm based on the on-line acquired information follows the geometrical-based initialization. The algorithm performance is evaluated in simulation and in particular the improvement on the solution brought by the Voronoi tessellation with respect to different initializations is analyzed.
\end{abstract}

\section{Introduction}
\begin{figure}[tb]
\centering
\includegraphics[width=.55\columnwidth]{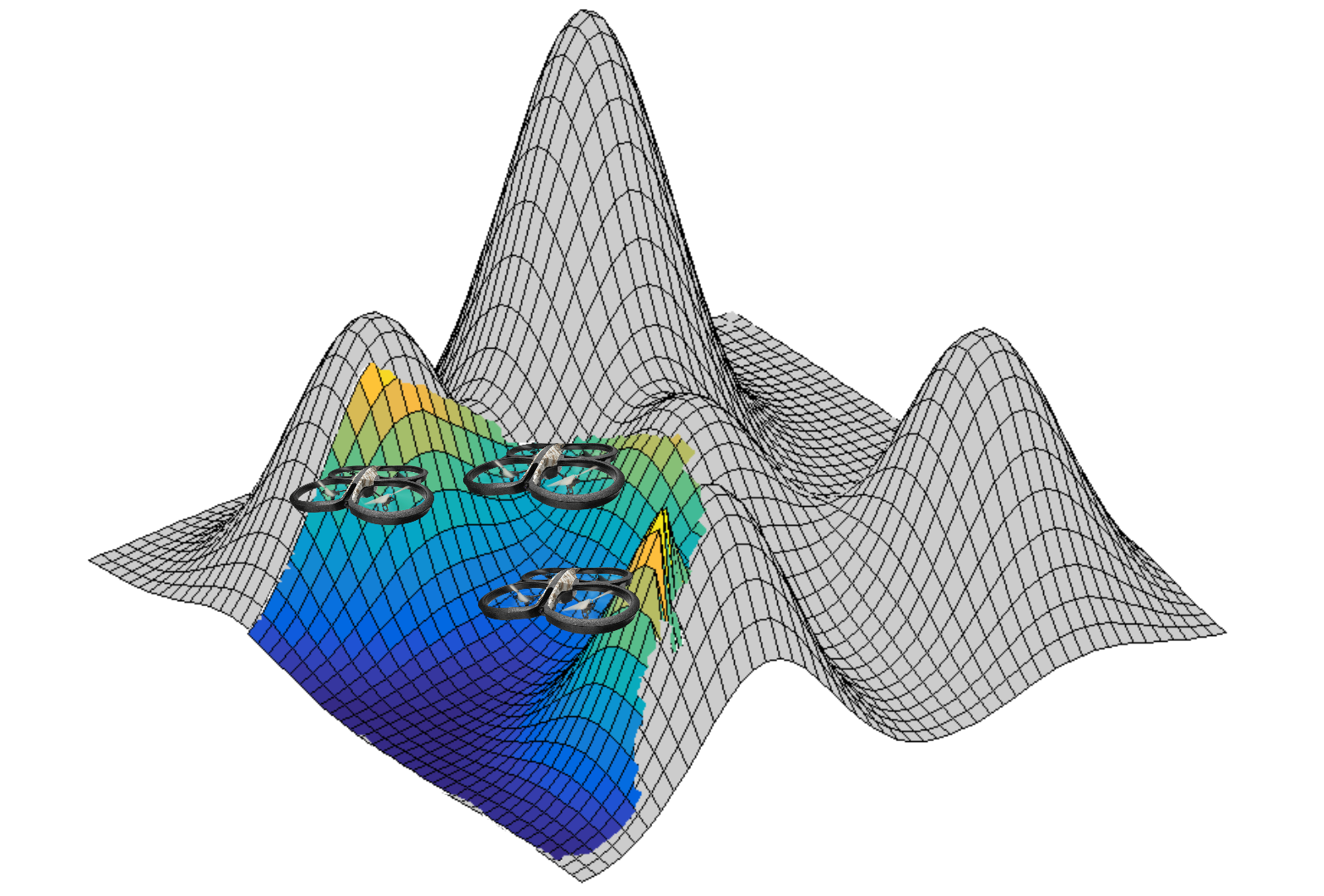}
\caption{Visual cooperative coverage: a team of UAVs has the goal to explore the environment to find the optimal deployment which maximizes the visible part of the surface.}
\label{fig:coverage}
\end{figure}

In the last years, it has been largely proved that the use of Unmanned Aerial Vehicles (UAVs) is an efficient and safe way to deploy visual sensor networks in complex environments. In this context, a widely studied problem is the cooperative coverage of a given environment. Among all the possible formulations of this problem, in this paper we focus on the visual blanket coverage, where the goal is to find the optimal static configuration of a team of UAVs equipped with a camera in order to maximize the visibility of a given three-dimensional region of interest (see Fig. \ref{fig:coverage}). This task is crucial for many important applications such as mapping and 3D reconstruction of rugged terrains and complex structures, environmental monitoring, surveillance and so on. In a typical scenario, a team of UAVs is called to achieve the mission without a perfect knowledge on the environment and needs to generate the trajectories on-line, only based on the information acquired during the mission through - usually noisy - measurements. For this reason, guaranteeing to obtain a global optimal solution of the problem is impossible in most cases. Furthermore, the presence of several constraints on the motion (collision avoidance, dynamics, etc.) as well as from limited energy and computational capabilities, makes this problem particularly challenging. A suitable way to tackle this problem is to adopt derivative-free optimization methods based on numerical approximations of the objective function, otherwise unknown in its exact analytic form. However, even though these methods allow finding a solution even when the environment is totally unknown and incorporating all the constraints of the problem, the final configuration can be strongly dependent on the initial positions and the system can get easily stuck in local optima very far from the global solution. A way to overcome this problem, common to every local optimization algorithm, can be found in initializing the optimization with a favorable configuration. In the cooperative visual coverage problem, the presence of an a priori partial knowledge on the environment to cover can be a fundamental source of information to exploit to this end. This kind of information is often available in many practical situations, where a rough knowledge of the shape or of few characteristic features are available or can be retrieved from the ground before the mission starts.


The main contribution of this paper is thus to study the effect of such initialization phase in these scenarios and in particular to propose a prepositioning of the UAVs by computing a constrained Centroidal Voronoi Tessellation \cite{okabe1992} on an approximating surface. In this way, a new layer is added to the optimization scheme in order to exploit as much as possible a prior partial and sparse information on the environment. For the second measurement-based local optimizer, we adopt a stochastic optimization algorithm recently presented for cooperative coverage problems \cite{renzaglia2012ijrr}. The resulting global method, taking advantages of the complementary properties of geometric and stochastic optimization, significantly improves the result of the previous approach and notably reduces the probability of a far-to-optimal final configuration. Moreover, the number of iterations necessary for the convergence of the on-line algorithm, and so the number of measurements and time required to accomplish the mission, is also reduced.

The rest of the paper is organized as follows. After a brief review of related work presented in Section \ref{sec:rel_work}, the problem tackled in this paper is formally defined in Section \ref{sec:prob_formulation}. Section \ref{sec:opt} presents the proposed optimization scheme adopted to obtain an optimal solution for the visual coverage problem, the main contribution of this work. To validate this approach, results obtained in simulated environments are presented and discussed in Section \ref{sec:results}.

\section{Related Work}
\label{sec:rel_work}

The problem of optimally covering a predefined area with a network of cooperating mobile sensors has been widely studied in recent years and several formulations of this problem have been proposed and analyzed. A first classification in three varieties of coverage was presented in \cite{gage1992coverage} where the concepts of blanket, barrier and sweep coverage were introduced. The focus of our work is on the blanket coverage, defined as the achievement of a static configuration which optimizes a certain coverage criterion. A fundamental work on this problem was presented by Cort\'{e}s {\it et al.} in \cite{cortes2004coverage}, where a solution based on the centroidal Voronoi partition was proposed to steer the agents toward an optimal configuration in two-dimensional convex areas. Successively, numerous papers presented extended versions of this approach considering more complex environments, different sensing models, etc. In this direction, an extension to the case of non-planar surfaces and including also non-convex environments is presented in \cite{breitenmoser2010}. The optimal visual coverage of a planar non-convex region is then considered in \cite{ganguli2006}, where the authors proposed a gradient-based algorithm for the case of a single robot and they prove that the visible area is almost everywhere a local Lipschitz function of the observer location. The same authors studied the multi-agent case in \cite{ganguli2007}, assuming omni-directional vision and unlimited sensing range, to achieve total visibility in orthogonal environments. A more complex sensing model is considered in \cite{papatheodorou2017}, where the authors studied the optimal visual coverage of a planar surface with a team of UAVs equipped with down-facing cameras, modeled as conical fields of view. A distributed control law maximizing a joint coverage-quality criterion is proposed, where the quality of the covered area is considered decreasing with the altitude. The case of visual optimal coverage for 3D terrains is approached in \cite{thanou2014distributed}, where the authors assume to have a complete knowledge of the environment to cover. Also in this case the solution is based on the concept of Voronoi tessellation of the environment. In \cite{zhang2015}, a different approach based on recursive convex optimization is presented for multi-camera deployment to visually cover a 3D object. However, also in this case, the authors assume a perfect initial knowledge of the object, specifically its 3D CAD model.

Most of the these approaches, based on geometric and gradient-based optimization algorithms, often requires to make several assumptions on the environment or on the constraints of the problem. Derivative-free approaches can instead be more suitable to tackle complex scenarios overcoming these limitations, as shown in \cite{renzaglia2012ijrr} where a stochastic optimization algorithm is adopted to guide a team of UAVs to optimally cover unknown 3D terrains. A modified version of this algorithm, including sparse regression techniques, has been also proposed in \cite{tseng2017near} where a target detection problem is tackled as an optimal coverage problem with the goal of finding minimum-time trajectories. The main drawback of this class of algorithms remains the strong dependency for certain scenarios on the initial conditions and their effect on the final solution. Our contribution precisely aims to significantly reduce this effect and so overcome this important limit.

\section{Problem Formulation}
\label{sec:prob_formulation}

Let us consider a team of $N$ Unmanned Aerial Vehicles called to cover an arbitrary three-dimensional surface $\mathcal{S}$, which for simplicity we consider rectangular along the $(x,y)$-axes, i.e. $x_{min}\leq x \leq x_{max}$, $y_{min}\leq y \leq y_{max}$. Let $P = \{ \mathbf{x}^{(i)}\}_{i =1,...,N}$ denote the configuration of the team, where $\mathbf{x}^{(i)}\in\mathbb{R}^3$ is the position of the $i$-th robot. We define a given point of the surface $\mathbf{q}\in\mathcal{S}$ visible from a team configuration if there exists at least one UAV for which:
\begin{enumerate}[(a)]
 \item the UAV and the point $\mathbf{q}$ are connected by a line-of-sight;
 \item the UAV and the point $\mathbf{q}$ are at a distance smaller than a given threshold value $d_{max}$ (sensing range).
\end{enumerate}
For a given configuration $P$, we define the visible surface $\mathcal{V}$ as the set of all $\mathbf{q}\in \mathcal{S}$ such that both the conditions (a), (b) are verified. The goal of the mission is to find the admissible UAVs configuration $P^*$ such that:
\begin{equation}
\label{eq:problem}
 P^* = \argmax_P \; \int_{\mathbf{q}\in\mathcal{V}(P)}\,\,d\mathbf{q} \,.
\end{equation}
Clearly, the visible surface $\mathcal{V}$, and so this integral, for arbitrary configurations $P$ cannot be computed at the beginning of the coverage mission, when the surface $\mathcal{S}$ is still unknown.

The surface of interest $\mathcal{S}$ is indeed considered arbitrary and unknown in its detailed structure except for a rough and limited information, represented by its $(x,y)$ limits and some sparse characteristic points. These can come from a rough knowledge of the environment, of its characteristics or even acquired from the ground before starting the task. This information, even though very limited, is important since used to construct an initial coarse approximation of the environment on which the first step of our optimization algorithm is based. In many realistic applications this knowledge is available or easy to retrieve and exploiting it as much as possible significantly improves a solution, otherwise based exclusively on the data acquired on-line with a blind start. In Section \ref{subsec:prior_info}, we will show how even a minimal information can be still exploited to this end.

Additionally, we require that the UAVs positions satisfy at every time a certain set of constraints so that robot-to-obstacle and robot-to-robot collision avoidance as well as minimum and maximum height of flight constraints are met. In other words, at each time-instant $k$, the vectors $\mathbf{x}_k^{(i)},i=1,\ldots, N$ must satisfy a set of constraints which, in general, can be represented as follows:
\begin{equation}
\label{eq:constr}
{\mathcal C}\left(\mathbf{x}_k^{(1)}, \ldots, \mathbf{x}_k^{(N)}\right) \leq 0 \,.
\end{equation}
Also in this case, the function $\mathcal{C}$ will depend on the surface and so its explicit expression is not available.

\section{Optimization scheme}
\label{sec:opt}

In this section we describe the two steps of the proposed optimization scheme to obtain an optimal configuration which maximizes the visibility of the environment: {\it i)} the off-line computation of the centroidal Voronoi tessellation generated by the robots positions on a surface approximating the real environment and {\it ii)} a constrained stochastic optimization algorithm based on on-line acquired information to allow the agents to converge to a local optimum of the problem.

\begin{figure}[tb]
\centering
\includegraphics[width=.55\columnwidth]{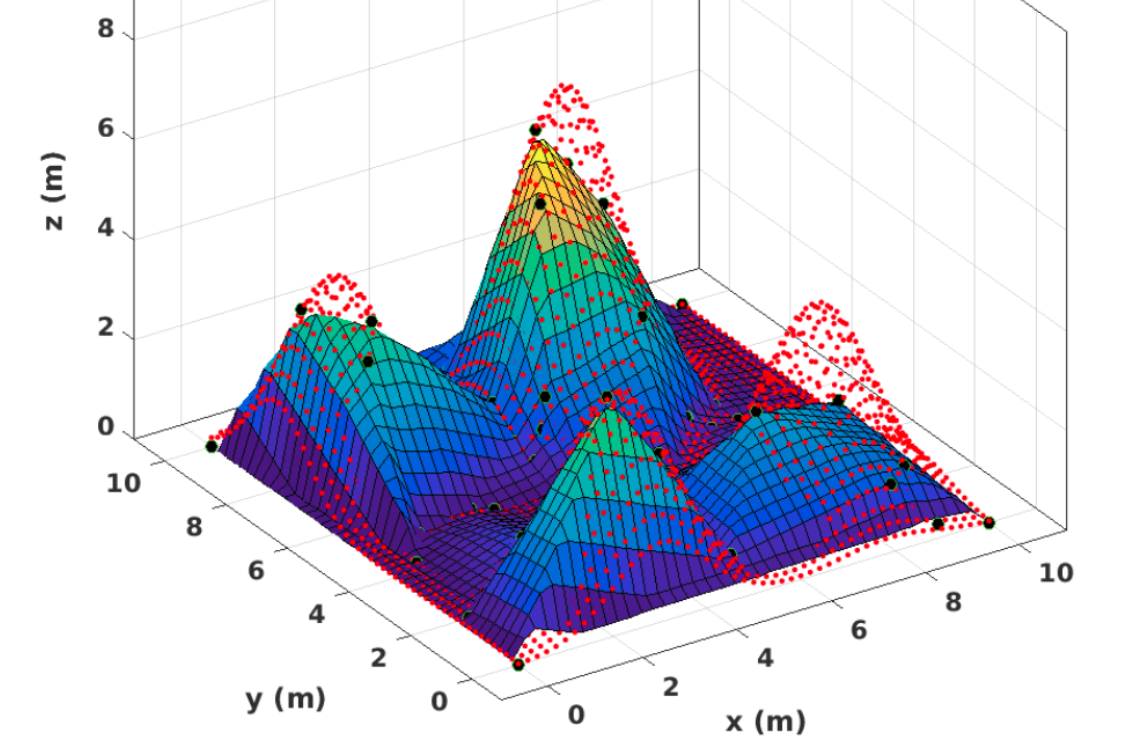}
\caption{Shape approximation: the real surface to cover (red point cloud) is approximated via a piecewise linear approximation based on a set of 50 randomly selected samples (black points).}
\label{fig:surf_approximation}
\end{figure}

As mentioned, we assume that a certain number of real or predicted points of the environment are available before starting the mission. Based on this initial information, standard methods for surface reconstruction starting from a point cloud can be applied to obtain an approximating surface $\mathcal{S}'$ for the real region to cover $\mathcal{S}$. This surface $\mathcal{S}'$ is then used as input for our optimization process. For simplicity sake, in this work we consider surfaces described as continuous functions of $(x,y)$ and we use a piecewise linear interpolation of the set of available points $\{\mathbf{q}_i^*\}_{i=1}^n$ as regression method. For more information on piecewise linear approximation of functions of two variables see e.g. \cite{piecewise2010} and references therein. Fig.~\ref{fig:surf_approximation} presents an example of an approximation, based on 50 randomly selected samples. It is worth to remark that more complex 3D structures would not affect the rest of the approach but would require a more complex reconstruction method.


Once an approximating surface is available, the following step of our approach is to compute a partition of this surface to have a first sub-optimal assignment of the UAVs, which can be used as an initialization of the successive local optimization. To partition $\mathcal{S}'$ we propose to compute the constrained centroidal Voronoi tessellation generated by the UAVs' positions.

\subsection{Voronoi Tessellation for Initial Positioning}
\label{subsec:voronoi}

A fundamental concept in Locational Optimization theory is the Voronoi tessellation \cite{okabe1992}. Given a bounded set $\Omega \in \mathbb{R}^n$ and a set of $N$ points
$\{\mathbf{z}_i\}_{i=1}^N$, the Voronoi tessellation of $\Omega$ generated by $\{\mathbf{z}_i\}_{i=1}^N$ is $\{V_i\}_{i=1}^N$, where

\begin{small}
\begin{equation}
 V_i = \{\mathbf{u}\in\Omega : |\mathbf{u}-\mathbf{z_i}|<|\mathbf{u}-\mathbf{z_j}|\;\forall j\neq i\},\, i=1,\ldots,N \,.
\end{equation}
\end{small}
Given a density function $f(\mathbf{x})\geq0$, the centroid $\mathbf{z}_i^*$ of the $i$-th region is defined as:
\begin{equation}
 \mathbf{z}_i^* = \frac{\int_{V_i}\mathbf{u}f(\mathbf{u})d\mathbf{u}}{\int_{V_i}f(\mathbf{u})d\mathbf{u}} \quad \text{for}\;i=1,\ldots,N \,.
\end{equation}
The tessellation is called Centroidal Voronoi Tessellation (CVT) if and only if $\mathbf{z}_i=\mathbf{z}_i^*, \forall i =1,\ldots,N$.

\begin{figure}[tb]
\centering
\includegraphics[width=.4\columnwidth]{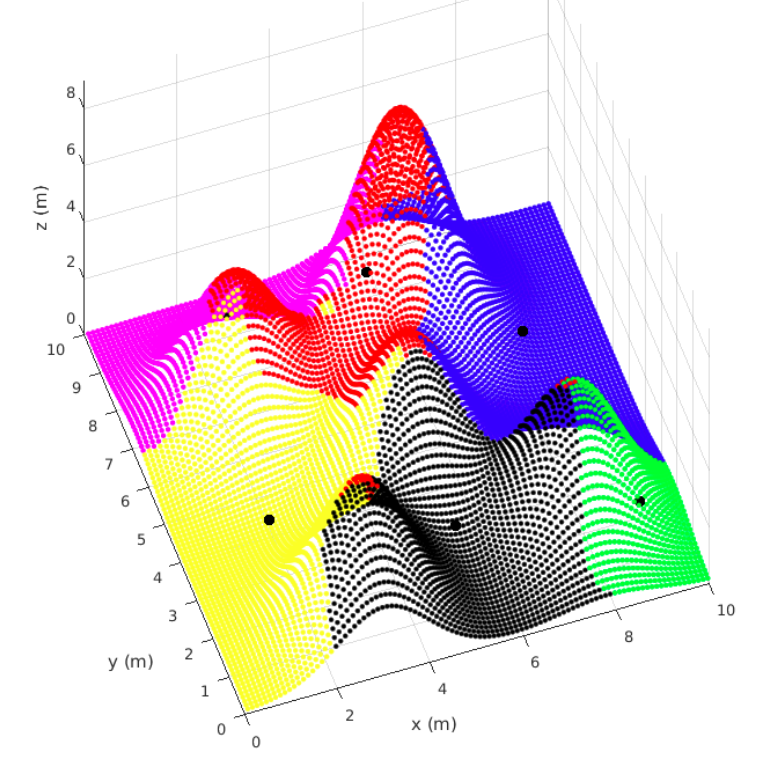}
\caption{Constrained CVT on a surface generated as a mixture of Gaussians with six generator points. Each region is represented with a different color.}
\label{fig:voronoi}
\end{figure}

As discussed in Section \ref{sec:rel_work}, CVT have been largely used as base to design optimal strategies for cooperative coverage problems, especially for 2D areas, and several algorithms have been developed to converge to a CVT starting from an arbitrary generators configuration. A commonly adopted solution is the Lloyd's algorithm \cite{lloyd1982, cortes2004coverage}, particularly efficient for simply planar domains but not suitable for more complex cases since it requires the actual computation of the Voronoi tessellation corresponding to the given set of points at every iteration. This task becomes extremely difficult for arbitrary surfaces in 3D and probabilistic algorithms are more suitable. In this paper we adopt a modified version of the MacQueen's method, initially proposed in \cite{ju2002probabilistic} and then extended to general surfaces in \cite{du2003voronoi}. At each iteration, the steps of this algorithm are: 
\begin{enumerate}
\item Fix a positive integer $q$ and constants $\{\alpha_i,\beta_i\}_{i=1}^2$ such that: $$\alpha_2>0,\, \beta_2>0,\, \alpha_1+\alpha_2=\beta_1+\beta_2=1\,;$$
\item Choose an initial set of $N$ random points $\{\mathbf{z}_{i=1}^N\}\in\mathcal{S}'$ and set $j_i=1$ for $i=1,\ldots,N$; 
\item Select randomly $q$ samples $\{\mathbf{y}_r\}_{r=1}^q \in \mathcal{S}'$;
\item For $r=1,\ldots,q$ determine $\mathbf{z}_{i_r^*}$ as the closest $\{\mathbf{z}\}_{i=1}^N$ to $\mathbf{y}_r$;
\item For $i=1,\ldots,N$ define the set $W_i$ as the set of all samplings $\mathbf{y}_r$ closest to $\mathbf{z}_i$ and compute the average $\mathbf{y}_i^*$ of the set $W_i$ and define:
\begin{equation}
 \hat{\mathbf{z}}_i^*  \leftarrow \frac{(\alpha_1 j_i+\beta_1)\mathbf{z}_i+(\alpha_2j_i+\beta_2)\mathbf{y}_i^*}{j_i+1}\,, j_i \leftarrow j_i+1 \,;
\end{equation}
the surface constraint is then imposed with a projection operator $\mathbf{z}_i^*=proj(\hat{\mathbf{z}}_i^*)$ and the set of $\{\mathbf{z}_i^*\}$ form the new set of points $\{\mathbf{z}_i\}_{i=1}^N$.
\end{enumerate}
These steps are iterated until a certain convergence criterion is fulfilled. Note that this algorithm does not require the calculation of Voronoi region on the surface. In Fig.~\ref{fig:voronoi} an example of constrained CVT with six generators is shown.


Note that such a configuration does not provide any guarantee in terms of visibility of the environment and thus does not represent an optimal solution of the problem. Its unique role is to produce a suitable preallocation which allows the successive optimization algorithm to significantly reduce the number of iterations required to converge and to overcome many local optima. It is worth remarking that an alternative method could have been to find a true global optimum over the approximated surface. However, the reason we do not consider this possibility is twofold: first, this step could be also performed on-line as a brick of a bigger mission and the computational time and capabilities necessary for a global optimization would be excessive for the limited resources usually available in these scenarios; second, such optimal solution would still not be a real optimum of the original problem since computed over a coarse approximation and not on the real surface.

\subsection{On-line Stochastic Optimization}
\label{subsec:stoch_opt}

The objective function (\ref{eq:problem}) we want to optimize is in practice a very complex function of the UAVs' positions and of the environment morphology, which we assume is not completely known. As a result, its explicit analytic expression is not available and a direct gradient-based algorithm cannot be employed. A different approach based on on-line gathered information is thus necessary. A viable solution in this case is represented by gradient-free stochastic optimization algorithms. These methods are indeed powerful tools that ensure convergence to a local optimum of the problem based on just noisy measurements collected during the process. However, especially in complex scenarios, the obtained local optimal solution can strongly depend on the algorithm initialization and be far from the global optimum. For this reason, the CVT plays a fundamental role in this approach.

In this paper, to obtain a solution to our problem, we consider as optimization method the Cognitive-based Adaptive Optimization (CAO) algorithm, which has been recently proposed to provide solutions for multi-robot deployment problems \cite{renzaglia2012ijrr}, \cite{sfly2014}. We provide here a description of this approach, however a more exhaustive presentation and analysis of its properties and performance can be found in the previous references. This method, originally developed and analyzed by Kosmatopoulos in \cite{kosmatopoulos2009cao}, represents an efficient way to solve in real-time constrained optimization problems where the exact forms of an objective functions ${\mathcal J}\left(\mathbf{x}_k^{(1)}, \ldots, \mathbf{x}_k^{(N)}\right)$ is not available. To circumvent this difficulty, the first step of the CAO approach is to make use of function approximators to obtain, at each iteration, a local estimation of this function ${\mathcal J}$.
The second step consists in selecting next agent positions such that the approximation function is maximized, respecting the constraints. These two steps are detailed below.\\
\noindent {\bf 1.}  At each time-instant $k$, ${\mathcal J}$ is estimated according to:
\begin{equation}
\label{eq:cao1}
\hat{J}_k\left(\mathbf{x}_k^{(1)}, \ldots, \mathbf{x}_k^{(N)}\right) = \vartheta_{k}^{T} \phi\left(\mathbf{x}_k^{(1)}, \ldots, \mathbf{x}_k^{(N)}\right) \,.
\end{equation}
Here $\hat{J}_k\left(\mathbf{x}_k^{(1)}, \ldots, \mathbf{x}_k^{(N)}\right)$ denotes the approximation of ${\mathcal J}$ generated at the $k$-th time-step, $\phi$ denotes the nonlinear vector of $L$ regression functions, $\vartheta_{k}$ denotes the vector of parameter estimates calculated at time $k$ and $L$ is a positive user-defined integer denoting the size of the function approximator (\ref{eq:cao1}). As regression functions we used polynomial terms up to third grade, whose terms are random combinations of the state variables. The parameter estimation vector $\vartheta_{k}$ is then calculated at each time $k$ based on a limited set of past measurements according to:
\begin{equation}
\label{eq:cao2}
\vartheta_{k} = \underset{\vartheta}{\argmin}\frac{1}{2} \sum_{\ell=\ell_{k}}^{k-1}\left(J^n_{\ell}- \vartheta^{T} \phi\left(\mathbf{x}_\ell^{(1)}, \ldots, \mathbf{x}_\ell^{(N)}\right) \right)^2
\end{equation}
where $\ell_{k}=\max\{0, k-L-T_h\}$ with $T_h$ being a user-defined non-negative integer. Standard least-squares optimization algorithms can be used to solve (\ref{eq:cao2}). Note that using only a limited set of past values has the advantage to significantly reduce the computation time required to obtain the parameter vector $\vartheta_k$ but the resulting function is a reliable approximation only locally and cannot be used for global optimization.\\
\noindent {\bf 2.} As soon as the estimator $\hat{J}_k$ is constructed according to (\ref{eq:cao1}) and (\ref{eq:cao2}), the new robots positions are obtained by directly testing on $\hat{J}_k$ sets of possible configurations generated by randomly perturbing the current state. Formally, a set of $M$ candidate state configurations is constructed according to:
\begin{equation}
\label{eq:pert}
\mathbf{x}_k^{i,j} = \mathbf{x}_k^{(i)} +\alpha_k \mathbf{\zeta}_k^{i,j}, i\in \{1, \ldots, N\}, j \in \{1, \ldots, M\} \,,
\end{equation}
where $\mathbf{\zeta}_k^{i,j}$ is a zero-mean, unity-variance random vector with dimension equal to the dimension of $\mathbf{x}_k^{(i)}$ and $\alpha_k$ is a positive real sequence which satisfies the standard conditions: 
\begin{equation}
\lim_{k\rightarrow\infty}\alpha_k=0, \quad \sum_{k=1}^\infty\alpha_k=\infty, \quad \sum_{k=1}^\infty\alpha_k^2<\infty \,.
\end{equation}
Among all $M$ candidate new configurations $\mathbf{x}_k^{1,j}, \ldots, \mathbf{x}_k^{M,j}$, the ones that correspond to non-feasible positions, i.e. the ones that violate the constraints (\ref{eq:constr}), are neglected and the new robots positions are calculated as follows:


\begin{equation}
\left[\mathbf{x}_{k+1}^{(1)}, \ldots, \mathbf{x}_{k+1}^{(N)}\right] = \underset{j \in \{1, \ldots, M\}} {\argmax} \hat{J}_k \left( \mathbf{x}_k^{1,j},\ldots, \mathbf{x}_k^{N,j}\right) \,.
\end{equation}

The random choice for the candidates is essential and crucial for the efficiency of the algorithm, because this choice guarantees that $\hat{J}_k$ is a reliable and accurate estimate for the unknown function ${\mathcal J}$. On the other hand, the choice of a slowly decaying sequence $\alpha_k$, a typical choice of adaptive gains in stochastic optimization algorithms (see e.g. \cite{bertsekas2000}), is essential for filtering out the effects of the noise term and ensuring final convergence. For more details on this optimization algorithm, choice of parameters and proof of convergence see \cite{renzaglia2012ijrr}.

\section{Simulation Results}
\label{sec:results}

\begin{figure*}[t]
\centering
\subfloat[]{\includegraphics[width=0.35\textwidth]{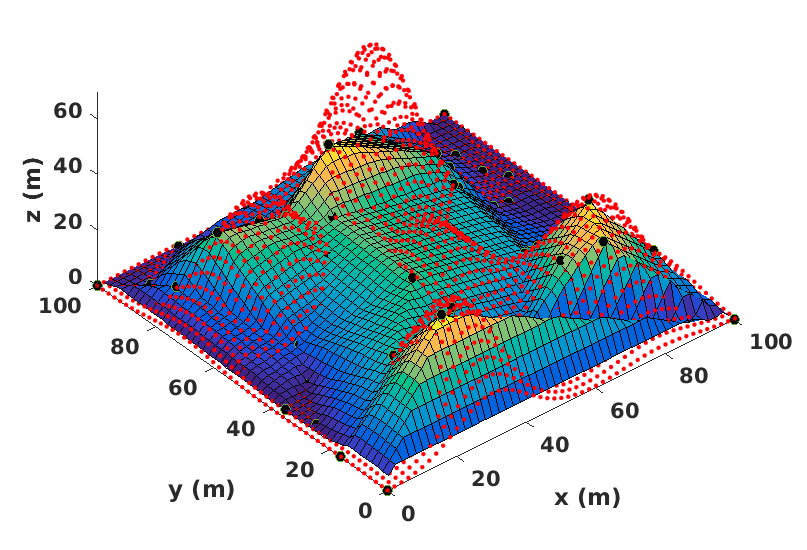}} \hspace{.5cm}
\subfloat[]{\includegraphics[width=0.35\textwidth]{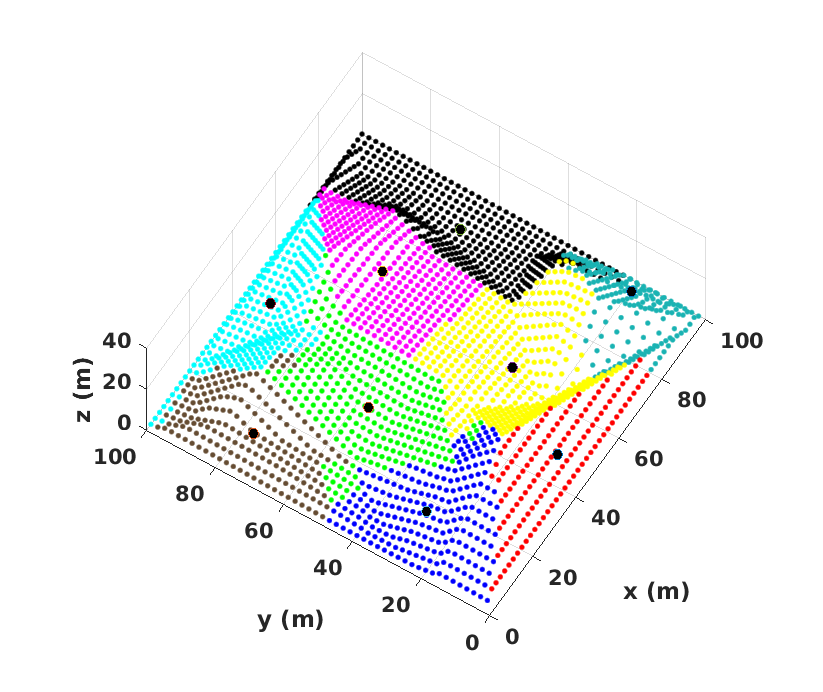}}\\
\subfloat[]{\includegraphics[width=0.36\textwidth]{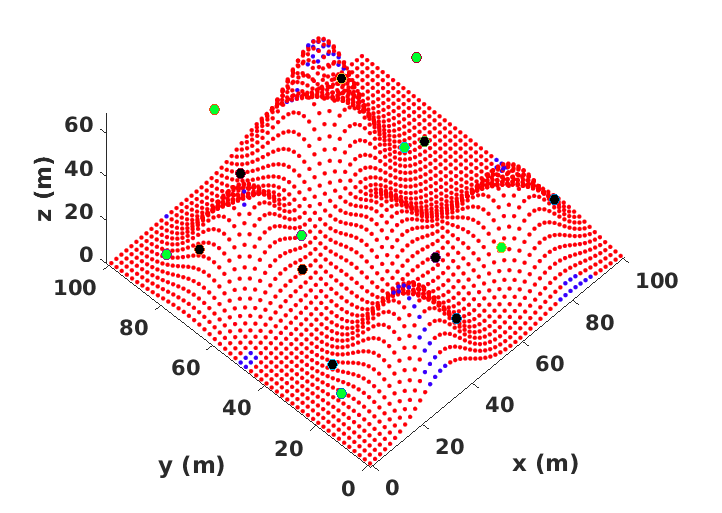}}  \hspace{1.5cm}
\subfloat[]{\includegraphics[width=0.3\textwidth]{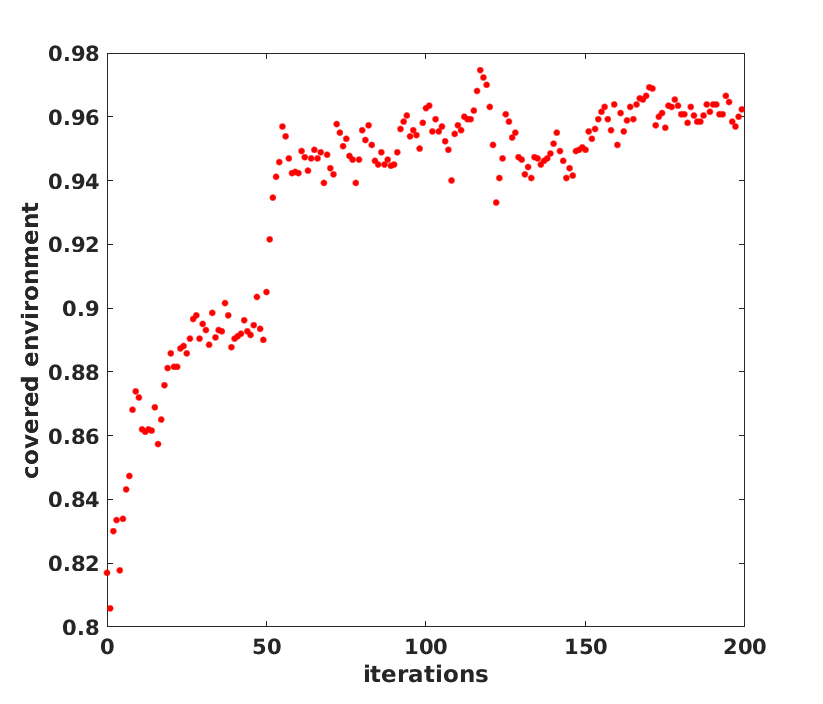}}
\caption{Coverage obtained with 9 robots and a sensing range $d_{max}=35m$. (a) The surface of interest (in red) is approximated by a piecewise linear interpolation of few sampling points (black dots). (b) Constrained CVT on the approximating surface. (c) The initial and final points are in black and green respectively. The red part of the surface is covered by at least one robot. (d) Portion of covered environment as function of CAO iterations.}
\label{fig:allsteps}
\end{figure*}

The proposed coverage algorithm has been tested in simulated environments with the main goal of showing the advantages of using the proposed CVT-based initial positioning as a way to overcome the actual limits of local optimization algorithms such as CAO.
To this end, in addition to present the performance of this approach for generic surfaces, comparative simulations using the same stochastic optimization algorithm but with a different or none initialization have been carried out. In particular, these comparisons aim to prove the importance of adopting the Voronoi-based initialization, even considering smooth surfaces, to achieve a faster convergence and to increase the robustness with respect to different environments and team capabilities. Furthermore, an analysis on the dependency of the final results on the quantity of information available on the surface to cover is also presented.

After quantitative results obtained in simplified environments to allow generating many random instances, a more realistic simulation, in terms of both UAVs model and environment to cover, has been carried out in Gazebo\footnote{http://gazebosim.org/} to show the employability of the approach in realistic scenarios.

\subsection{Experimental Settings}
Throughout our experiments, we assume the UAVs to have omni-directional sensing capabilities constrained by a maximum sensing distance, as described in Section \ref{sec:prob_formulation}. The trajectories are defined as a sequence of way points and we assume the presence of an internal control law allowing the UAVs to follow them. The environments to cover are assumed to be arbitrary and two different scenarios have been considered: $i)$ regular terrains generated as mixtures of Gaussians and $ii)$ a typical outdoor environment composed of some buildings. These regions are assumed to be initially unknown except for a set of points, which in simulation have been randomly selected over the surfaces. In both cases, the coverage function in eq. (\ref{eq:problem}) for a given team configuration has been computed by discretizing the environment and considering the number of points belonging to the surface visible from the UAVs. To facilitate the comparison of results corresponding to different environments, the values of coverage level have been always normalized with respect to the total surface to cover. The parameters of the CAO algorithm have been set as proposed by the authors of \cite{renzaglia2012ijrr}.

\subsection{Illustrative Example}

Before providing more quantitative results, we start our study  by presenting in Fig. \ref{fig:allsteps} an illustrative example where the main steps of our approach are shown. In this instance, a team of 9 UAVs with a maximum sensing range of 35$m$ has the goal of covering an environment for which a limited set of points (30 points randomly selected) is the only initial information available. The real surface to cover is so approximated by a piecewise linear interpolation based on this point set (Fig. \ref{fig:allsteps}(a)). The approximated surface is then used to compute the constrained CVT generated by the UAVs positions (Fig. \ref{fig:allsteps}(b)). This configuration serves as the initial state for the stochastic optimization algorithm which, via an incremental perturbation of these positions, finally converges to a local optimum of the constrained optimization problem (Fig. \ref{fig:allsteps}(c)). The behavior of the coverage level during this last optimization step is shown in Fig. \ref{fig:allsteps}(d), where it is possible to see that from the final configuration the team is able to monitor the environment almost entirely. It is important to note that the initial positions provided by the CVT can be below the real surface, or at least below the minimum safe height of flight, due to surface approximation errors. In this case, we assume the UAVs able to sense the distance to the ground and converge to the closest admissible positions avoiding any collision with the ground.

\subsection{Comparative Results}


\begin{figure}[tb]
\centering
\includegraphics[width=.65\columnwidth]{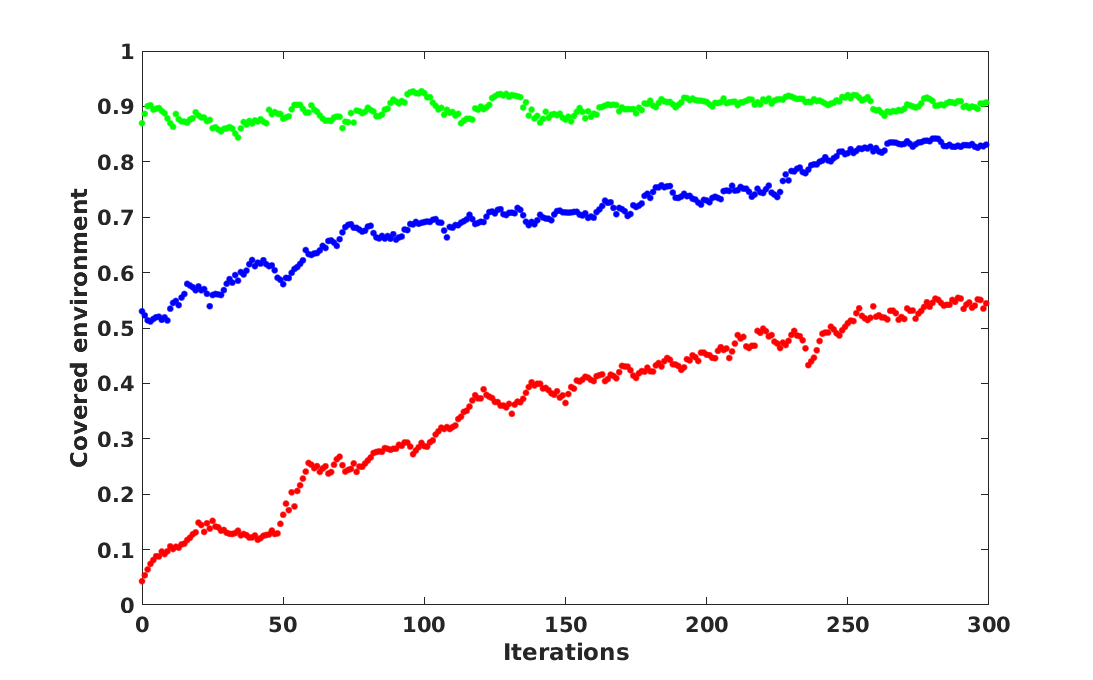}
\caption{Portion of covered environment during the optimization process. The team is composed of 10 robots. In red the robots start close one each other in a corner of the environment, in blue their positions are initialized randomly with a uniform distribution over the entire area and in green the Voronoi-based initialization is used as first step of the coverage approach. The maximum obtained values for the three strategies are 0.56, 0.84 and 0.93 respectively.}
\label{fig:comparison1}
\end{figure}

We remind that the benefit of including a CVT step to initialize the algorithm is mainly twofold: firstly, reducing the number of iterations for the on-line part of the algorithm required to reach a certain coverage level and secondly, even more important, finding a solution closer to the global optimum avoiding many local maxima of the problem. In order to show these two phenomena, we compare the results obtained by the proposed approach with the coverage levels achieved {\it i)} without any initialization, i.e. with all the UAVs starting with close positions in a corner of the environment, and {\it ii)} with randomly initialized positions generated by a uniform distribution over the entire region to cover.

We begin this analysis presenting, for a scenario similar to the one depicted in Fig. \ref{fig:allsteps}, the behavior of the coverage level achieved during the optimization process for the three different initializations. These results, presented in Fig. \ref{fig:comparison1}, show how the local optimization algorithm without any initialization is not able to find a good solution and the team remains stuck in a local optimum. A better solution can be obtained with an initial configuration already spread over the environment thanks to a random positioning of the resources. However, the Voronoi-based solution is able to start from a value of covered environment even higher than the final levels achieved in the other two cases. In this particular case, the initialization is already very close to a local optimum of the problem and the on-line algorithm can only slightly improve this initial result.

To have more statistically meaningful results, we present the same comparison but varying the number of robots participating in the task and the environment to cover. In particular, for each fixed number of UAVs and maximum sensing range, we consider 40 different scenarios. Each scenario is generated as a mixture of seven Gaussians centered in randomly selected locations, with varying height and fixed variance. For the results using the Voronoi-based initialization, a set of 30 randomly selected points is used to approximate the surface. We studied four different team sizes, varying from a minimum of 5 up to 20 UAVs and all the simulations are bounded with 500 iterations, which ensure the convergence of the optimization algorithm. The average values of maximum visual coverage achieved and respective standard deviations $\sigma$ for each case are reported in Table~\ref{tab:results1}. Additionally, when a non-trivial initialization is present, also the average initial values and their standard deviations are reported.

These results show that the algorithm with the Voronoi-based initialization always reaches the highest portion of covered environment and with the smallest standard deviation, showing a better robustness with respect to changes in the environment than the other approaches. To be more specific, the CVT initialization produces an improvement with respect to the random initialization between a minimum of 5$\%$ and a maximum of 12$\%$. Besides the maximum values obtained, comparing the coverage levels after the initialization is also interesting. In this case it is even more clear the advantage with respect to a random initialization, which also translates in a significant reduction of time to accomplish the mission.

It is also worth to briefly analyze the behavior of these performances with respect to variations in the team size. On this subject, a more thorough study on the results obtained with the CAO algorithm was previously conducted in \cite{renzaglia2012ijrr}, showing also the ability of the on-line optimization to scale up to larger teams. Similarly, our experiments show that the proposed approach can easily handle large teams, as in the case of a team of 20 UAVs. We can also notice that the improvement obtained with the Voronoi-based initialization is more significant in the intermediate cases, i.e. with 8 and 12 UAVs. This is motivated by the fact that they represent more challenging scenarios where the optimal configuration is harder to find. With few UAVs, it is indeed sufficient to simply obtain a good spread within the team to reach configurations with non-intersecting fields of view. In such configurations the team is then close to its maximum sensing capability and so they correspond to close-to-optimal solutions. On the other side, in very numerous teams a high level of overlap between fields of view is inevitable and several different configurations can potentially lead to the same total visibility of the environment (in the limit case of a very large number of UAVs, even a random deployment might lead to a complete coverage of the environment). However, in this case, an initial dispersion of the robots can still be important to avoid that some of them remain stuck in their initial position reducing the team sensing capability. This effect can be seen in the case with 20 UAVs, where the result without initialization is significantly lower than the other two values, which are both close to a complete coverage.

\begin{table*}[ht]
\centering
\footnotesize
    \renewcommand{\arraystretch}{1.2}
    \vspace{.2cm}
 \begin{tabular}{|| c || c | c || c | c || c | c || c | c || c | c ||}
 \hline
 & \multicolumn{4}{|c||}{\textbf{CVT initialization}}& \multicolumn{4}{|c||}{\textbf{Random initialization}}& \multicolumn{2}{|c||}{\textbf{No initialization}}\\
 \hline
 Number of robots & Initial & $\sigma$ & \;\;Max.\;\; & $\sigma$ & Initial & $\sigma$ & \;\;Max.\;\; & $\sigma$ & \;\;Max.\;\; & $\sigma$ \\
 \hline\hline
 5 & 0.59 & 0.06 & {\bf 0.65} & 0.04 & 0.37 & 0.07 & 0.61 & 0.05 & 0.57 & 0.09 \\ 
 \hline
 8 & 0.74 & 0.04 & {\bf 0.84} & 0.03 & 0.52 & 0.08 & 0.75 & 0.04 & 0.62 & 0.09 \\
 \hline
 12 & 0.84 & 0.02 & {\bf 0.92} & 0.02 & 0.63 & 0.07 & 0.86 & 0.03 & 0.69 & 0.08 \\
 \hline
 20 & 0.90 & 0.02 & {\bf 0.98} & 0.01 & 0.73 & 0.05 & 0.93 & 0.02 & 0.76 & 0.08 \\
 \hline
\end{tabular}
\vspace{.2cm}
 \caption{Average and standard deviation of coverage levels achieved over 40 scenarios corresponding to randomly generated terrains in a 100x100$m^2$ square for various numbers of UAVs. The sensing range $d_{max}$ is fixed to 25$m$. In case of an initialization step, also the initial coverage level is reported.}
 \label{tab:results1}
\end{table*}

\begin{figure*}[tb]
\centering
\includegraphics[trim=.5cm 1.6cm 0cm 0cm,width=0.32\textwidth]{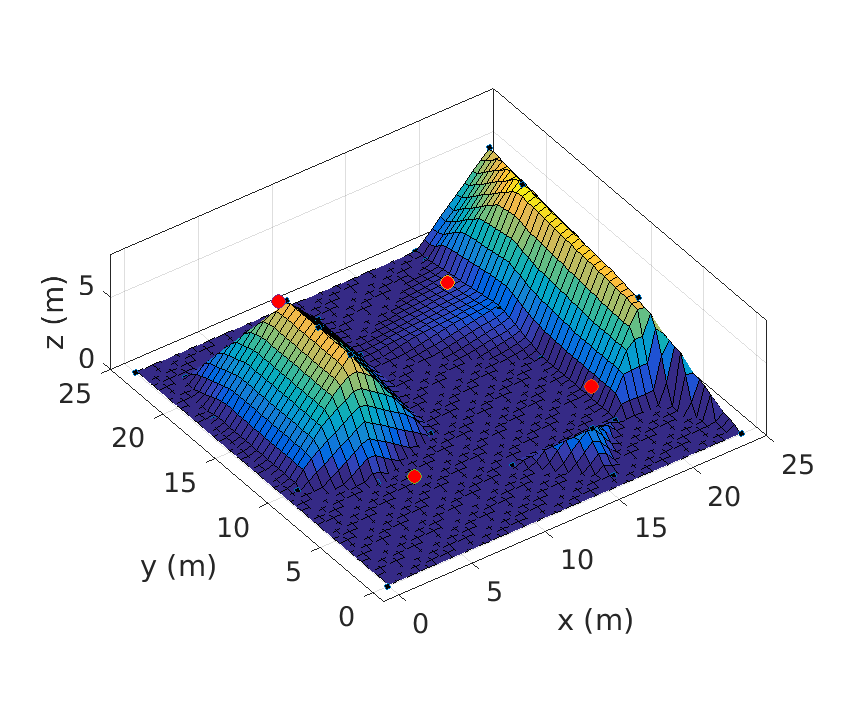}\hspace{-.1cm}
\includegraphics[trim=0cm 0cm 0cm 0cm,width=0.3\textwidth]{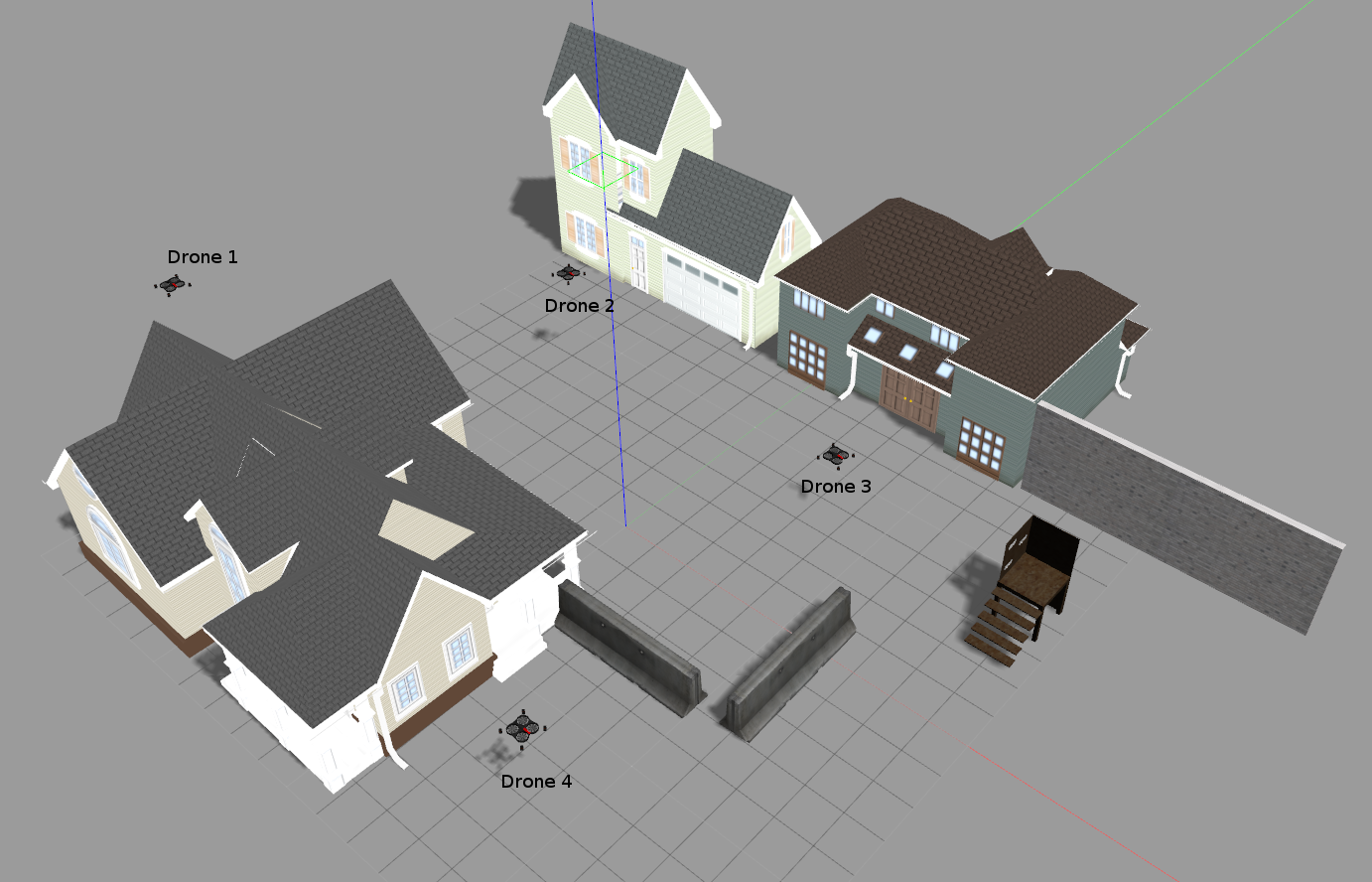}\hspace{.2cm}
\includegraphics[trim=0cm 0cm 0.5cm 0cm,width=0.31\textwidth]{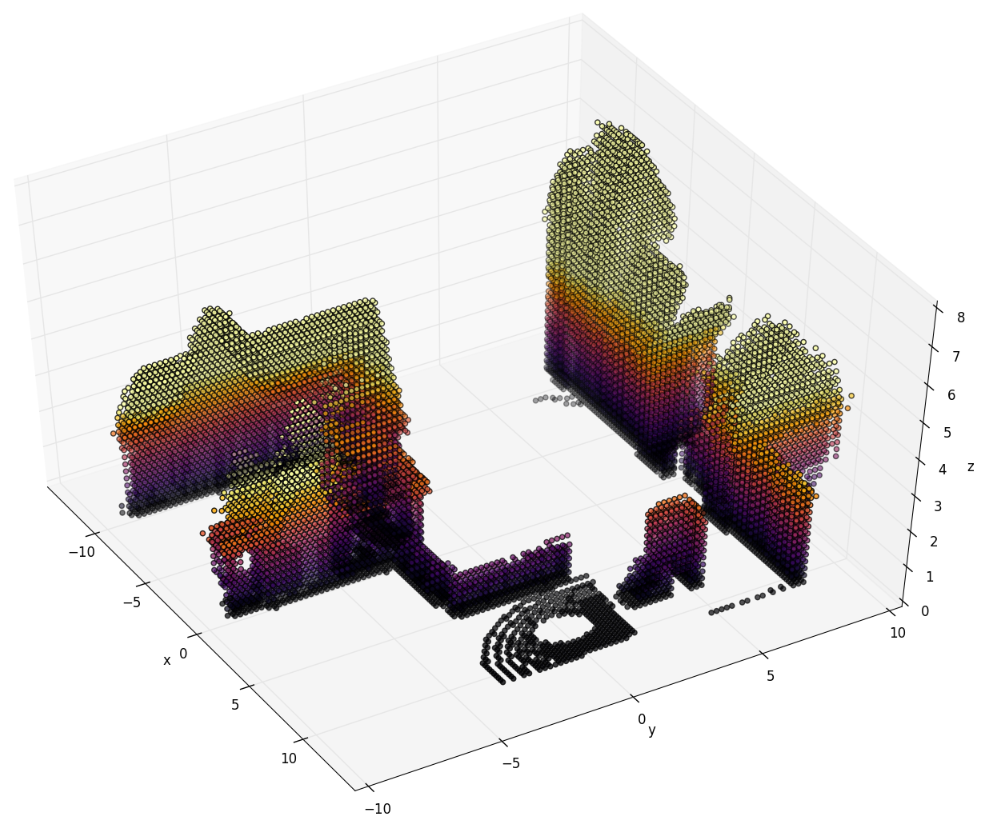}
\vspace{-0cm}\caption{Four UAVs covering a simulated environment with several buildings. From left to right: an approximation of the environment allows computing the initial positions, the CAO algorithm optimizes them on-line in the actual scenario leading to the final configuration and the point cloud reconstructing the environment obtained by the team.}
\label{fig:gazebo_sims}
\end{figure*}

\subsection{Dependency on Prior Information}
\label{subsec:prior_info}
As a final result, we want to remark the fact that the CVT allows the team to uniformly distribute over the environment and reach a good initialization even if the information on the real surface is extremely limited. For one scenario previously analyzed (12 robots, third line in Table~\ref{tab:results1}), we carry out the same study over the forty environments but considering the approximating surface obtained with just the four points describing the boundaries of the region of interest plus one single point in the middle of the environment. In this way the resulting approximation is simply a pyramid. The coverage obtained after the initialization with this input surface is $0.78$ with a standard deviation $\sigma=0.03$, while the final maximum coverage achieved is $0.91$ with a standard deviation $\sigma=0.02$. The first result shows that, not surprisingly, the initial value is lower than the one obtained by exploiting a more informative approximation, but it still remains a better solution than a random initialization. Then, from this configuration, the optimization algorithm is able to lead the system to reach a final coverage performance very close to the value of the more informative surface, almost vanishing the initial advantages carried by the latter. This clearly highlights the important properties of the Voronoi tessellation to distribute the robots and provide a suitable initialization.


\begin{figure}[tb]
\centering
\includegraphics[width=.65\columnwidth]{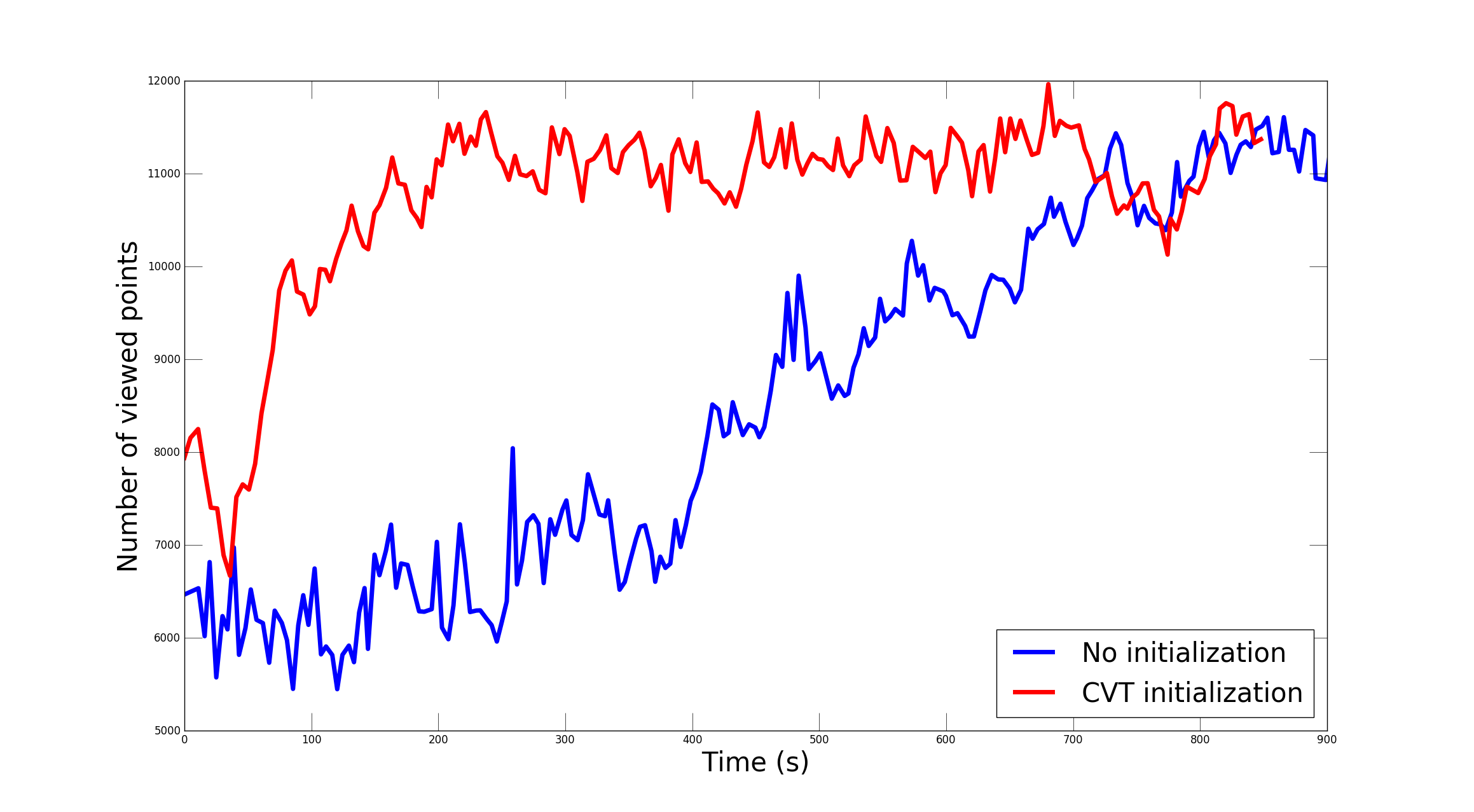}
\caption{Coverage levels for the Gazebo scenario presented in Fig. \ref{fig:gazebo_sims}, corresponding to the CVT initialization and to a non-initialized configuration.}
\label{fig:gazebo_coverage}
\end{figure}

\subsection{Realistic Environment}
We present here the results obtained considering more realistic models for both the environment and the system to show the feasibility and employability of our approach in real applications. The simulations have been carried out in Gazebo and realized with the ROS package {\it tum\_simulator}\footnote{http://wiki.ros.org/tum\_simulator}, which includes a simulator of the Parrot AR drone 2.0. In the proposed scenario, four simulated AR drones with a maximum visual range of $8m$ are employed to cover a $24$x$24m^2$ area with several buildings. Fig.~\ref{fig:gazebo_sims} presents the scenario and the obtained results: on the left, the initialization is computed on a very simplified map of the environment; these positions are then optimized on-line by the CAO algorithm based on measurements of visibility taken in the actual environment (middle image); on the right, the environment covered by the team seen as a point cloud. Finally the coverage level achieved during the mission is shown in Fig. \ref{fig:gazebo_coverage}, where it is also compared with the result obtained from a different initialization, i.e. two UAVs starting close to each other on one side of the environment and the other two similarly on the opposite side. It is clear that the initialized case begins from a higher values and converge significantly faster to the final positions - not far from the initialization - showing the importance of starting the optimization from preallocated positions.

\section{Conclusion}
\label{sec:conclusion}
In this paper we showed the importance of exploiting prior knowledge on the environment to achieve optimal multi-UAV visual coverage. In particular, we proposed a CVT (Centroidal Voronoi Tessellation) based initialization of agents on an approximating surface, in order to improve the performance of stochastic optimization algorithms with no or poor initialization.
The presence of such initialization significantly reduces the probability of being trapped in local optima far from the global optimal solution. At the same time, the number of iterations, and thus of measurements and computation, required for the algorithm to converge is also lower. This advantage can be crucial for the success of missions in complex environments where small UAVs with limited energy are employed. Results obtained in simulated experiments showed the improvement carried by this two-step scheme with respect to the optimization algorithm without any or with a random initialization.

In the future, it is our intention to extend our study to different strategies to escape local optimality in these scenarios, considering advantages and drawbacks in terms of energy consumption and final performance. Also an analysis of more informative coverage criteria, including the quality of the acquired visual data in the objective function, will be another interesting direction. This could also affect the generation of the initial partition since new metrics could be considered instead of standard distances for the Voronoi tessellation. Finally, we aim to test this approach in more realistic scenarios, reconstructed from real data. This would lead us toward a final implementation on a team UAVs for coverage missions in real outdoor environments.


\bibliographystyle{IEEEtran}
\bibliography{biblio}

\end{document}